\pdfoutput=1

\documentclass[11pt]{article}

\usepackage[preprint]{acl}

\usepackage{times}
\usepackage{latexsym}

\usepackage[T1]{fontenc}

\usepackage[utf8]{inputenc}

\usepackage{microtype}

\usepackage{inconsolata}

\usepackage{graphicx}

\usepackage{booktabs}
\usepackage{amsmath}
\usepackage{booktabs}
\usepackage[table]{xcolor}  
\usepackage[np]{numprint}
\usepackage{colortbl}
\usepackage{multirow}
\usepackage[font=small, labelfont=bf]{caption}
\usepackage{url}
\usepackage[normalem]{ulem}
\usepackage{xcolor}

\usepackage{hyperref}

\newcommand{\posneg}[1]{%
  \begingroup
  \ifdim #1pt<0pt
    \color{red}%
  \else
    \color{green!50!black}%
  \fi
  \ensuremath{#1}%
  \endgroup
}

\definecolor{lightgray}{gray}{0.9}

\title{Surprisal and Metaphor Novelty Judgments: Moderate Correlations and Divergent Scaling Effects Revealed by Corpus-Based and Synthetic Datasets}

\author{Omar Momen, \quad Emilie Sitter,  \\
\textbf{Berenike Herrmann} and \textbf{Sina Zarrieß} \\
     CRC 1646 – Linguistic Creativity in Communication \\  Faculty of Linguistics and Literary Studies \\ Bielefeld University, Germany \\
      \small{
\texttt{\{omar.hassan,emilie.sitter,berenike.herrmann,sina.zarriess\}@uni-bielefeld.de}}}

\begin{document}
\maketitle
\begin{abstract}
Novel metaphor comprehension involves complex semantic processes and linguistic creativity, making it an interesting task for studying language models (LMs). This study investigates whether surprisal, a probabilistic measure of predictability in LMs, correlates with annotations of metaphor novelty in different datasets. We analyse the surprisal of metaphoric words in corpus-based and synthetic metaphor datasets using 16 causal LM variants. We propose a cloze-style surprisal method that conditions on full-sentence context. Results show that LM surprisal yields significant moderate correlations with scores/labels of metaphor novelty. We further identify divergent scaling patterns: on corpus-based data, correlation strength decreases with model size (inverse scaling effect), whereas on synthetic data it increases (quality–power hypothesis). We conclude that while surprisal can partially account for annotations of metaphor novelty, it remains limited as a metric of linguistic creativity.\footnote{Code and data are publicly available: \url{https://github.com/OmarMomen14/surprisal-metaphor-novelty}.}
\end{abstract}

\section{Introduction}

\begin{figure}[!htp]
  \centering
  \includegraphics[width=1\linewidth]{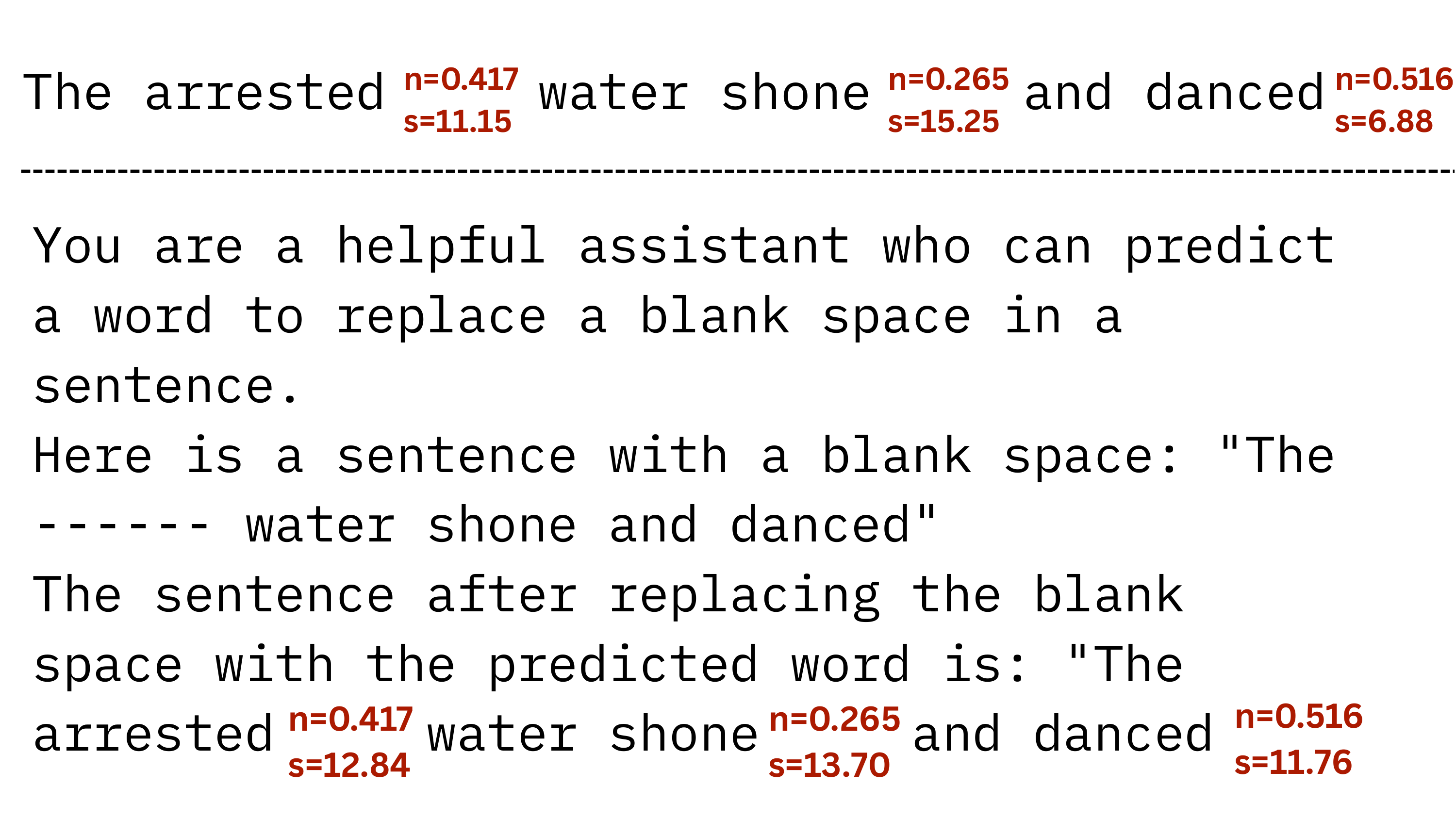}
  \caption{A sentence from VUA-ratings with metaphor-novelty ratings (n) and surprisal measures (s) from GPT2-base. The top part shows the direct-surprisal measure, while the bottom part shows the cloze-surprisal measure. 
  }
  \label{fig: nov_surp_example}
\end{figure}

Recent advances in language modelling have led to a renewed interest in studying linguistic creativity, an aspect of language that was notoriously challenging for traditional NLP systems.
A well-known instance of creative language is metaphor, which arises through mapping of a source domain–a lexical unit’s literal meaning–onto a target domain–its figurative meaning-\cite{cmt1980}.
Yet, not all metaphors are equally novel or creative as many of such mappings are highly conventionalized, such as in the famous example ``to \textit{attack} an argument,'' where the mapping from the source domain ``WAR'' onto the target domain ``ARGUMENT'' corresponds to a conventional sense of the verb ``attack'' which can be found in dictionaries. Other metaphors, as in ``The \textit{arrested} water'' (Figure \ref{fig: nov_surp_example}) establish a more unconventional or novel mapping that is considered creative.

The distinction between conventional and novel metaphors is well established in theory~\cite{BowdleGentner2005,Gibbs_2006} and studied in experimental work on cognitive processing~\cite{ArzouanGoldsteinFaust2007, lai2009}.
Novel metaphors require greater interpretative effort compared to conventional metaphors, as the unfamiliar mapping requires speakers to construct new connections between domains~\cite{philip_2016_conventional_novel}.

However, separating novel metaphors from conventional ones remains a challenging task for human annotators~\cite{do-dinh-etal-2018-weeding,reimann-scheffler-2024-metaphor}.
While computational methods on metaphor analysis do not often take the novelty dimension into account~\citep{maudslay-teufel-2022-metaphorical}, and some existing studies highlight that novel metaphors are more difficult to detect than conventional ones~\cite{neidlein-etal-2020-analysis,tong-etal-2024-metaphor}.
In this work, we propose to fill this gap by studying questions related to LMs' processing of novel and conventional metaphors and investigating correlations between LM-based metrics and different setups of metaphor novelty annotations.

We draw on a line of research that has studied the ability of LMs to account for effects of difficulty in human sentence processing and that goes back to surprisal theory \citep{hale-2001-probabilistic,levy2008expectation}.
Surprisal is computed with LMs as the negative log probability of a word in context and has been found to provide a robust predictor for human processing difficulty (e.g., of reading times) 
\citep{goodkind-bicknell-2018-predictive,shain2024large}.

However, recent work draws a mixed picture in terms of which LMs can provide the most robust and cognitively plausible predictors for processing difficulty. \citet{oh-schuler-2023-surprisal} observe an inverse scaling effect when testing surprisal estimates from GPT-2 models of different sizes, showing that surprisal computed with smaller model sizes achieved a better fit with human reading times than larger models. \citet{wilcox-etal-2023-language}, on the other hand, train LMs of small and medium size on a range of languages and find that LM quality generally correlates with its predictive power of reading times.

In this paper, we investigate surprisal of metaphoric words computed using a selection of LMs as a metric of metaphor novelty. We find significant, positive and moderate correlations between LMs' surprisal and different metaphor novelty annotations. 
We perform our experiments on four different datasets coming from corpus-based and synthetic setups, using LMs of different sizes. Most interestingly, we observe effects supporting the "inverse scale" pattern~\cite{oh-schuler-2023-surprisal} on the two corpus-based novelty datasets, and contrary patterns supporting the counter-argument, "Quality-Power" hypothesis~\cite{wilcox-etal-2023-language} on the two synthetic datasets. We also investigate the effect of instruction-tuning on surprisal correlation to metaphor novelty. Moreover, we conduct a deeper analysis of the genre splits, revealing that genre, metaphor density, and LM perplexity are potential factors underlying the quality of surprisal as a predictor of metaphor novelty. Finally, we introduce a new method of computing surprisal, \textit{cloze-surprisal}, to include the right context of the word in its conditional probability. We find that this method can boost the correlation of surprisal with metaphor novelty annotations by a few points. In general, our study establishes a promising direction for studying linguistic creativity with LMs and calls for novel measures and datasets that provide systematic annotations of metaphor novelty across genres.

\section{Related Work}
\label{sec: related work}
\subsection{Metaphor Annotations}
The most common annotation scheme in corpus-based metaphor studies is the Metaphor Identification Procedure Vrije Universiteit (MIPVU)~\cite{mip2007,vuamc}, designed as a reliable step-by-step framework for identifying metaphorically used words. This method was used to construct the VU Amsterdam Metaphor Corpus (VUAMC)~\cite{vuamc}, a large-scale, genre-balanced corpus containing 186,673 words sampled from fiction, news, academic writing, and conversations splits in the BNC Baby edition. The VUAMC has become a benchmark for metaphor research, inspiring developments in further English corpora~\cite{beigman-klebanov-etal-2018-corpus,mohammad-etal-2016-metaphor} and more recent multilingual efforts~\cite{sanchez-bayona-agerri-2022-leveraging,egg-kordoni-2022-metaphor}.

Several studies have extended annotation beyond binary literal/metaphoric labels. 
\citet{mohler-etal-2016-introducing} introduced a four-point metaphoricity scale, 
considering factors such as vividness and familiarity. \citet{Reijnierse2018} proposed the concept of \textit{deliberateness}, distinguishing metaphors intended to be recognized as metaphors; novel metaphors are typically deliberate under this framework. Direct novelty annotations were first introduced by \cite{parde_corpus_2018}, who asked annotators to rate metaphorical word pairs from the VUAMC on a 0–3 novelty scale. \citet{do-dinh-etal-2018-weeding} proposed a more comprehensive approach considering all metaphor words in VUAMC, and by aggregating ranked novelty judgments into continuous scores. More recently, \citet{reimann-scheffler-2024-metaphor} proposed a dictionary-based method that labels a metaphor as novel when its contextual meaning is absent from dictionary entries.

Psycholinguistic studies also produce datasets of metaphors, but these are smaller in size and feature controlled synthetic sentences (fixed words, structures, sentence length, etc.), cf. \cite{cardillo2010neural,cardillo2012selective,roncero2014role}.
Some of these also created datasets of metaphors that are classified as conventional or novel metaphors \cite{lai2009,ahrens-burgers-zhong-2024}.

\subsection{Predictive Powers of Surprisal}

Surprisal estimates from LMs have been shown to capture human processing difficulty across multiple behavioural and neural measures. In self-paced reading and eye-tracking studies, token-level surprisals from causal models significantly predict reading times beyond lexical and syntactic factors~\cite{goodkind-bicknell-2018-predictive,wilcox2020,oh-schuler-2023-transformer}. Similar effects have been reported for acceptability judgments, where LM surprisals align with human sensitivity to grammaticality~\cite{lau-etal-2017,meister-etal-2021-revisiting,hu-etal-2020-systematic,tjuatja-etal-2025-goes}.

\subsection{LM-Based Metrics for Metaphor Novelty}

Recent work has explored diverse LM-based approaches—prompting, surprisal, embedding similarity, and attention patterns—for the analysis of metaphorical language. \citet{aghazadeh-etal-2022-metaphors} demonstrate that metaphor-relevant information is encoded in mid-layer embeddings of multilingual pre-trained LMs. \citet{ichien2024largelanguagemodeldisplays} show that GPT-4 generates
interpretations of novel literary metaphors that are favoured by human judges over interpretations by college students, suggesting sensitivity to metaphorical meaning beyond lexical overlap. 
\citet{pedinotti-etal-2021-howling} tested BERT’s masked token probabilities across conventional, novel, and nonsensical metaphors, showing that novel metaphors tend to receive lower probabilities, but distinctions between novelty and nonsense remained unclear. We extend this study towards more recent LMs and broader comparisons between models, datasets, and genres. \citet{djokic2021episodic} trained a BERT-based classifier to predict novelty scores jointly with the task of metaphor detection.

\section{Metaphor Novelty Datasets 
}
\label{sec: met_nov_ann}

This Section describes the four datasets we use in our experiments and explains how metaphor novelty is annotated in different approaches. Datasets statistics are reported in Tables~\ref{tab: data_stats} and~\ref{tab: data_stats2} (Appendix~\ref{sec: ap_stats_examples}).

\subsection{Corpus‐based Datasets}
\label{subsec:corpusbased}
Following the MIPVU procedure, all words in the VUAMC are annotated as being a metaphor-related word (MRW) or not. Out of 186,673 words in VUAMC, 24,762 words (15,155 content words) are labelled as MRWs. While the original VUAMC does not include novelty annotations, we utilise two annotation studies that offer different annotation protocols for metaphor novelty on the same corpus.

\paragraph{VUA-ratings} \citet{do-dinh-etal-2018-weeding} collected crowd-sourced ratings of metaphor novelty for VUAMC. In their set-up, annotators were presented four sentences containing an MRW (content words only) and asked to select the best (most novel) and worst (most conventional).
Each MRW appeared in six different best-worst scaling comparisons. These annotations were then converted into continuous Best-Worst Scaling scores~\cite{kiritchenko-mohammad-2017-best} in the range of (-1, +1), with -1 being the most conventional and +1 being the most novel. Additionally, they also convert these scores to binary labels using a threshold of 0.5. This results in labelling 353 metaphors as novel out of 15,155 content metaphoric words in VUA.

\paragraph{VUA-dictionary} \citet{reimann-scheffler-2024-metaphor} proposed a dictionary-based method that labels a metaphor as novel when its contextual meaning is absent from dictionary entries. They applied this method to VUAMC. In particular, they re-annotated the potentially novel metaphors in VUA according to VUA-ratings and~\citet{Reijnierse2018}'s metaphor deliberateness annotations (1,160 potentially novel metaphors in total). Their procedure resulted in labelling only 409 content\footnote{We exclude non-content words from this dataset to allow comparison with VUA-ratings, which originally excluded non-content words.} metaphoric words as novel out of the 1,160 potentially novel metaphors. We assume the remaining metaphors in VUA to be conventional in our study.

\subsection{Synthetic Datasets}
\label{subsec:synthetic}

Another class of metaphor novelty datasets is synthetic datasets, which are mainly characterised by being generated from a fixed source. They usually have comparable sentences in terms of the target metaphoric words. We consider two cases of such datasets, a dataset from a psycholinguistic study concerned with novel metaphors. And a toy dataset that we generated from GPT-4o.

\paragraph{Lai2009} \citet{lai2009} investigated how our brains handle conventional and novel metaphors differently. To this end, they constructed a set of controlled experimental items featuring metaphors with two degrees of novelty. Two linguists selected 104 words and constructed 4 sentences for each word, according to the Conceptual Metaphor Theory (CMT)~\cite{cmt1980}. For each word, the items include one (i) literal use, (i) conventional metaphor, (iii) novel metaphor, and (iv) anomalous use, with the target word as the last word in each sentence. Familiarity and interpretability tests showed a significant difference between the conventional and novel metaphoric senses. For our experiments, we select only the conventional and novel metaphor senses for each word, resulting in 208 sentences (104 conventional and 104  novel metaphors).


\paragraph{GPT-4o-metaphors} 
To test our experiments in a setting that is more controlled (in contrast to VUA) but features more varied sentence lengths, structures and degrees of novelty (in contrast to Lai2009), we construct a synthetic dataset by prompting GPT-4o to generate sentences that include conventional and novel metaphoric senses. We do not assume that LLMs are able to generate ideal novel metaphors, or that this dataset is a benchmark of any kind; we are only interested in exploring a new quick setting of potential metaphor novelty annotations. 
We prompt GPT-4o to generate 5 verbs and 5 nouns that can be used in a metaphoric sense.  
For each word, we prompt the model again to generate 10 different sentences using the target word in a conventional metaphor sense, and 10 different sentences in a novel metaphor sense. This results in a dataset of 200 sentences, 100 contain conventional metaphors, and 100 contain novel metaphors.

\section{Methods \& Experiments}
\label{sec: methods}

We describe our experiments designed to investigate whether and to what extent surprisal scores computed from LMs correlate with different annotations of metaphor novelty.

\subsection{Surprisal of a target word}

In information theory~\cite{shannon1948}, the information content of an event \(x\) with probability \(p(x)\) is defined as: $I(x) = -\log p(x)$. In the context of LMs trained to predict the next token in a sequence,
this quantity is called surprisal and computed for a $w_i$ in a sequence such that: \( \mathrm{Surprisal}(w_i) = -\log p(w_i \mid w_{<i}) \)\footnote{We use $\log$ of base $e$ for all surprisal/perplexity computations in our study.}.

In our experiments, we measure word-level surprisal of metaphoric words in their sentence-level context. We feed every sentence to an LM in an independent teacher-forced forward pass, and record surprisal of the target (metaphoric) word(s). We denote this quantity as \textit{direct-surprisal} to distinguish it from \textit{cloze-surprisal}.

Most LMs operate on subwords rather than words, so deriving word-level probabilities requires implementation choices that can affect surprisal estimates and, in turn, downstream measures~\cite{pimentel-meister-2024-compute,lesci-etal-2025-causal}. For transparency and reproducibility, we report two choices in our implementation that affect the computed surprisal values.
\textbf{First},
we choose to align tokens (subwords) to words by precomputing the offsets of the exact\footnote{We make sure no punctuation or any other characters outside of the word boundary are included inside the offsets.} target words in their corresponding sentences, and then searching for the minimal span of tokens in the LM's tokenisation of the input sentence covering the target word's offsets. This usually results in a proper alignment with the target word in addition to a leading whitespace character (e.g. \texttt{``Ġarrested''} for ``arrested''), with very rare cases when the last token is attached to a punctuation (e.g. \texttt{[``Ġindivid'', ``ual'', ``ism,'']} for ``individualism''). We also apply the corrections made by~\cite{pimentel-meister-2024-compute,oh-schuler-2024-leading} that address the problem of leading whitespaces in tokenisations of most causal LMs. \textbf{Second}, surprisal is computed using the conditional probability of the current word being the target word given the preceding words in a sentence. This makes computing surprisal of the first word in a sentence an issue\footnote{ In some cases in the VUA datasets, the target word is the first word in the sentence.}. To compute surprisal of the first word in a sentence, we prepend the input sentence with a "beginning of sequence" special token.

\paragraph{Cloze-surprisal}
A key limitation of standard (left-to-right) surprisal as a proxy for semantic contextual properties (e.g., metaphor novelty) is that it does not condition on future (right) context in the sentence. This matters in naturally occurring corpus data (e.g VUA datasets), where metaphorical words can appear in many sentence positions, and in some cases in the GPT-4o-metaphors dataset (see Table~\ref{tab: data_examples}). To incorporate the right context while retaining an autoregressive setup, we compute \emph{cloze-surprisal}: for each target word, we prompt the model to predict a missing word in a sentence, and we replace the target in the sentence by a blank space, and then we append the same sentence again as the intended completion (Figure~\ref{fig: nov_surp_example}). We measure surprisal for the target word at its position in the second occurrence of the sentence. Because the first occurrence exposes the full right context, the resulting conditional probability effectively incorporates both left and right contexts of the target word.

\subsection{Evaluating the correlation}

To determine whether LMs' surprisal correlates with different metaphor novelty annotations, we use multiple correlation metrics between surprisal and metaphor novelty scores/labels. 
For continuous novelty scores (as in VUA-ratings), we compute \textbf{Pearson's $r$}~\cite{pearson1895regression} and \textbf{Spearman's $\rho$}~\cite{spearman1904association} correlations.
For binary labels (conventional vs. novel), we compute 
the \textbf{Rank-biserial $r_b$}~\cite{glass1966rankbiserial} correlation. Rank-biserial estimates the probability that a random observation from the set of novel metaphors has a larger surprisal than one from the set of conventional metaphors, minus the reverse probability.
We also estimate the potential of surprisal as a discriminator for binary novelty labels using the Area Under the ROC Curve \textbf{(AUC)}~\cite{fawcett2006roc}. We also compute the significance of all these estimates.
All our metrics and tests do not assume normality, except for Pearson's, for which we also provide a non-parametric alternative (Spearman).

\subsection{Settings}

Surprisal measures are derived from the learnt $p(x)$ of pre-trained LMs and, hence, depend substantially on the model architecture and the training data. 
Also, human annotations of metaphor novelty depend on the annotation process. 
Additionally, novelty norms may vary by genre: a metaphor considered novel in academic texts might appear conventional in fiction. We thus investigate how all these factors affect the correlation by experimenting with multiple settings.

\paragraph{Models}

We examine three families of decoder-only causal LMs (GPT-2, Llama 3 and Qwen2.5)~\cite{gpt2, llama3, qwen2}. For each model family, we select 3-4 different sizes, to represent the effect of model size on the correlation. To investigate the effect of instruction-tuning on the correlation, we include the instruction-tuned variants of Llama 3 and Qwen2.5.

\paragraph{Datasets}

As explained in Section \ref{sec: met_nov_ann}, we perform experiments 
on four datasets (Tables~\ref{tab: data_stats} and~\ref{tab: data_stats2}). As VUA-ratings has continuous novelty scores, we evaluate the correlation of its continuous scores to surprisal. In addition, we convert these scores to binary labels using a threshold of $0.5$ and evaluate their correlation to surprisal, allowing us to compare the results of VUA-ratings to VUA-dictionary, Lai2009 and GPT-4o-metaphors, which only have binary novelty annotations.

\paragraph{Genre variables} VUA provides genre splits (see Section \ref{subsec:corpusbased}), allowing us to analyse the correlation between surprisal and novelty separately for each genre.

\paragraph{Perplexity} Scales of surprisal can differ from one genre to another, and from one model to another. To this end, we investigate perplexity as a potential factor in our study. Perplexity of a model on a certain dataset is the exponential of the average token-level surprisal of all tokens in the dataset. In that sense, perplexity indicates how much a certain model is ``surprised'' on average when predicting a certain dataset. We measure models' perplexity on the genre splits of VUA by feeding sentences one by one to the model, measuring token-level surprisals accordingly and averaging and exponentiating them to obtain perplexity. This yields higher values of perplexity than common values reported in literature, as we use a shorter context (single sentences).

\subsection{Summary}

Our experiments rely on a large collection of surprisal measures for each metaphor word in the four datasets: 32 different surprisal values (including direct and cloze) for 15,155 metaphoric words of the VUA dataset and for 208 and 200 metaphoric words of Lai2009 and GPT-4o-metaphors, respectively. 
In Figure~\ref{fig: scores_surp_hist} (Appendix~\ref{sec: apx_dists}), we plot the distributions of surprisal values by model, in addition to metaphor novelty scores/labels from VUA. From these plots and further normal distribution tests, we find that novelty scores and surprisal values are not strictly normally distributed.

\section{Results}
\label{sec: results}

We report the correlation measures for the corpus-based datasets in Table~\ref{tab: results_corpus-based}, and for synthetic datasets in Table~\ref{tab: results_synthetic}. We also report the gains (in terms of rank-biserial) of the instruction-tuned variants over their base models in Table~\ref{tab: results_instruct_gain}. And the gains of the cloze-surprisal method over the direct-surprisal method in Table~\ref{tab: results_cloze_gain}.

\subsection{Overall Correlation Results}

Generally, we find LMs' direct surprisal values correlate positively with metaphor novelty annotations across the four datasets (Tables~\ref{tab: results_corpus-based} and~\ref{tab: results_synthetic}). All correlation estimates are statistically significant for both the corpus-based and synthetic datasets~\footnote{The relatively large corpus-based datasets (15,155 datapoints) can bias the significance values; however, we get similar significance values for the relatively small synthetic datasets (100 datapoints).}. Overall, for direct surprisal, the largest Pearson correlation $r=.419$, largest Spearman correlation $\rho=.417$, largest rank-biserial correlation $r_b=.638$ and largest $AUC=.819$ come from GPT2-base on VUA-ratings. These values indicate a significant positive correlation, yet its strength is moderate.

We also find that the correlation strengths' ranges differ across the different annotation datasets. By comparing the rank-biserial estimate $r_b$ across the four datasets, we find its ranges as follows: (.47-.64) for VUA-ratings, (.43-.58) for VUA-dictionary, (.28-.50) for Lai2009, and (.38-.63) for GPT-4o-metaphors. These results show that surprisal correlates more strongly with the corpus-based data than with the synthetic data.
Also, surprisal correlates more strongly with human ratings on VUA than the dictionary-based annotation approach. While the most controlled dataset (Lai2009) gets the weakest correlations with surprisal.

\begin{table}
\centering
\small
\begin{tabular}{l | c c c c | c c}
\toprule
\textbf{Model} & \multicolumn{4}{c}{\textbf{VUA-ratings}} & \multicolumn{2}{|c}{\textbf{VUA-dict.}} \\
\midrule
& \textbf{$r$ }& \textbf{$\rho$} & \textbf{$r_b$} & \textbf{$auc$} & \textbf{$r_b$} & \textbf{$auc$} \\
\midrule
GPT2-base	&	.419	&	.417	&	.638	&	.819	&	.581	&	.791 \\
GPT2-med	&	.389	&	.383	&	.600	&	.800	& .557	&	.778 \\
GPT2-large	&	.381	&	.373	&	.585	&	.793	& .539	&	.769 \\
GPT2-xl	&	.373	&	.362	&	.566	&	.783	& .539	&	.769	\\
\arrayrulecolor{black!30}\midrule
\arrayrulecolor{black}
Llama-1B	&	.345	&	.329	&	.532	&	.766	&	.480	&	.740	\\
Llama-3B	&	.328	&	.308	&	.502	&	.751	&	.446	&	.723	\\
Llama-8B	&	.314	&	.293	&	.488	&	.744	&	.431	&	.716	\\
\arrayrulecolor{black!30}\midrule
\arrayrulecolor{black}
Qwen-0.5B	&	.384	&	.377	&	.598	&	.799	&	.543	&	.771	\\
Qwen-7B	&	.334	&	.314	&	.502	&	.751	&	.456	&	.728	\\
Qwen-14B	&	.316	&	.295	&	.470	&	.735	&	.430	&	.715	\\
\bottomrule
\end{tabular}
\caption{Correlation estimates between surprisal and novelty scores/labels in the corpus-based datasets. All results are statistically significant at the 0.001 level.}
\label{tab: results_corpus-based}
\end{table}

\begin{table}
\centering
\small
\begin{tabular}{l | c c | c c}
\toprule
\textbf{Model} & \multicolumn{2}{c}{\textbf{Lai2009}} & \multicolumn{2}{|c}{\textbf{GPT-4o-met.}} \\
\midrule
& \textbf{$r_b$} & \textbf{$auc$} & \textbf{$r_b$} & \textbf{$auc$}  \\
\midrule
GPT2-base	&	.276	&	.638	&	.511	&	.756	\\
GPT2-med	&	.362	&	.681	&	.586	&	.793	\\
GPT2-large	&	.397	&	.699	&	.629	&	.814	\\
GPT2-xl	&	.414	&	.707	&	.587	&	.794	\\
\arrayrulecolor{black!30}\midrule
\arrayrulecolor{black}
Llama-1B	&	.450	&	.725	&	.508	&	.754	\\
Llama-3B	&	.451	&	.725	&	.511	&	.755	\\
Llama-8B	&	.483	&	.742	&	.557	&	.778	\\
\arrayrulecolor{black!30}\midrule
\arrayrulecolor{black}
Qwen-0.5B	&	.374	&	.687	&	.382	&	.691	\\
Qwen-7B	    &	.494	&	.747	&	.469	&	.734	\\
Qwen-14B	&	.504	&	.752	&	.536	&	.768	\\
\bottomrule
\end{tabular}
\caption{Correlation estimates between surprisal and novelty labels in the synthetic datasets. All results are statistically significant at the 0.001 level.}
\label{tab: results_synthetic}
\end{table}

\begin{table}
\centering
\small
\begin{tabular}{l | c c c c}
\toprule
\textbf{Model} & \textbf{VUA-r} & \textbf{VUA-d} & \textbf{Lai} & \textbf{GPT-4o}  \\
\midrule
Llama-1B-It.      & \posneg{+4.0}  & \posneg{+1.5} & \posneg{+0.4} & \posneg{-5.1} \\
Llama-3B-It.      & \posneg{+4.0}  & \posneg{+2.4} & \posneg{-5.8} & \posneg{-5.6} \\
Llama-8B-It.      & \posneg{+1.6} & \posneg{+0.6} & \posneg{-2.3} & \posneg{-5.1} \\
Qwen-0.5B-It.     & \posneg{-1.6} & \posneg{-0.8} & \posneg{+0.1} & \posneg{+2.2} \\
Qwen-7B-It.       & \posneg{-1.7} & \posneg{-1.9} & \posneg{-8.9} & \posneg{-13.7} \\
Qwen-14B-It.      & \posneg{-3.9} & \posneg{-3.1} & \posneg{-3.7} & \posneg{-14.0}  \\
\bottomrule
\end{tabular}
\caption{Instruction-tuning \% gains (over corresponding base variants) in Rank-biserial's correlation estimates between surprisal and novelty scores/labels in the four datasets: \textbf{VUA-r}atings, \textbf{VUA-d}ictionary, \textbf{Lai}2009 and \textbf{GPT-4o}-metaphors.}
\label{tab: results_instruct_gain}
\end{table}

\subsection{Model Size Effects}

\begin{figure*}[!htp]
  \centering
  \includegraphics[width=\linewidth]{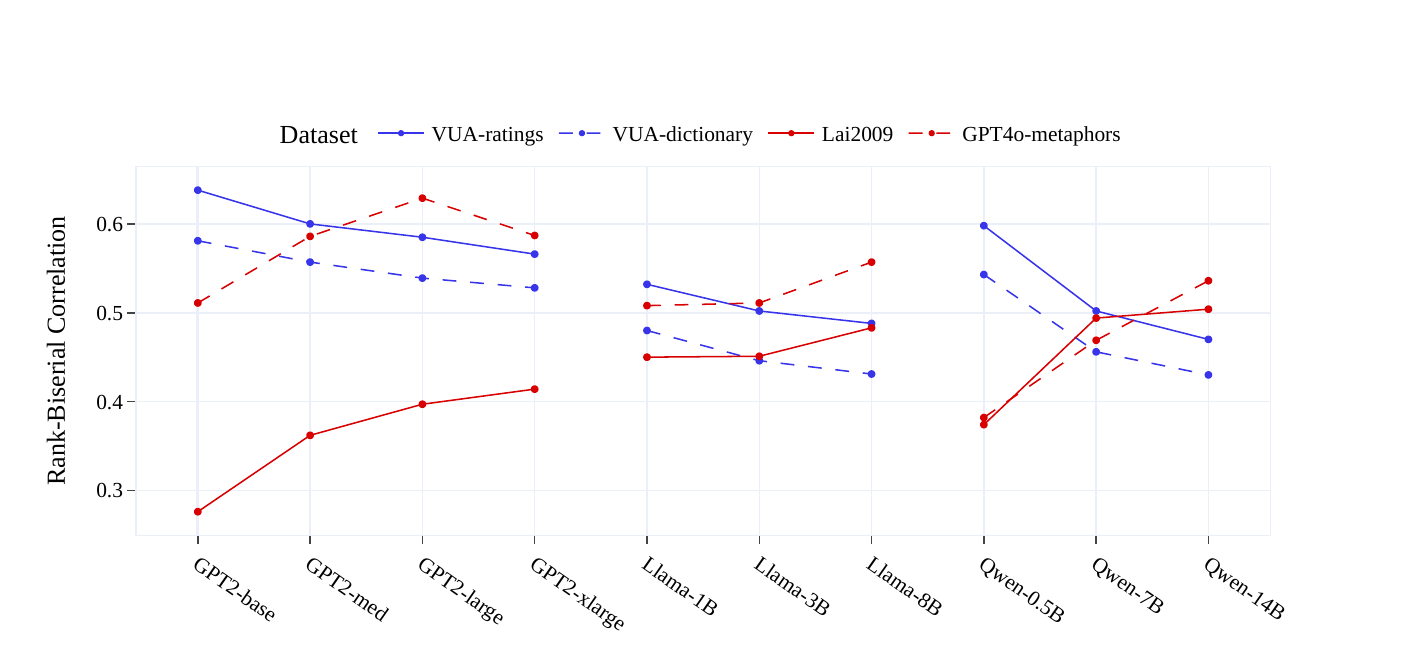}
  \caption{Effect of model size on correlation between surprisal and metaphor novelty annotations from the four datasets. Corpus-based datasets (blue) show a negative scale effect, while synthetic datasets (red) show a positive scale effect.}
  \label{fig: model_size effect}
\end{figure*}

The effect of model size is consistent (but diverging in direction) across the two dataset types. We plot the rank-biserial correlations of the ten model variants for the four datasets in Figure~\ref{fig: model_size effect}. For the two corpus-based datasets (blue), surprisal--novelty correlation is monotonically decreasing as model size increases (per model family). On the other hand, for the two synthetic datasets (red), surprisal--novelty correlation increases as model size increases, with a single minor exception at the GPT2 model family on the GPT-4o-metaphors dataset.

\subsection{Instruction-tuning Effects}

We report percentage gains in rank-biserial correlation for the instruction-tuned variants over their base variants across the four datasets in Table~\ref{tab: results_instruct_gain}. This shows the effect of extracting surprisals from an instruction-tuned variant over a base variant with the same experimental setting. We do not add a prompt/instruction to the instruction-tuned models' inputs.
The results suggest that instruction-tuning does not always improve the correlation between surprisal and novelty annotations. Only Llama instruction-tuned variants could improve the correlations over their base variants on the VUA datasets; otherwise, instruction-tuning deteriorates the correlations over the base variants. Also, instruction-tuning fails to improve the correlations over more basic models such as GPT2-base.

\subsection{Cloze-surprisal}
In Table~\ref{tab: results_cloze_gain}, we report the percentage gains in rank-biserial correlations for cloze-surprisal over direct-surprisal from same model variants across the four datasets. We find cloze-surprisal to be boosting the correlations with extra points in many cases; however, it also deteriorates the correlations heavily in other cases. 
For the GPT2 model family, cloze-surprisal consistently improves the corpus-based datasets' correlations, yielding the strongest correlations in this study. 
Out of the 32 recorded surprisals on each dataset, cloze surprisal from GPT2-base achieves the highest correlations on VUA-ratings, with $r=.499$, $\rho=.499$, $r_b=.687$ and $auc=.843$. However, cloze-surprisal of GPT2 models consistently deteriorates the correlations 
on the synthetic datasets.
Llama and Qwen model families also show positive gains from cloze surprisal in most of the cases, however, not as consistently as for GPT2 models on VUA. Most importantly, unlike the GPT2 model family, Llama and Qwen boost correlations on the synthetic datasets except for a very few cases. Cloze-surprisal achieves the overall strongest correlation on the Lai2009 dataset using the Qwen-14B model, with $r_b=.539$ and $auc=.779$. Also, it achieves the overall strongest correlation on the GPT-4o-metaphors dataset using the Llama-8B model, with $r_b=.684$ and $auc=.842$.

To investigate the effect of incorporating the whole sentence context in surprisal computation, we plot the distribution of LMs' direct and cloze surprisals of metaphoric words in VUA in Figure~\ref{fig: scores_surp_hist} (Appendix~\ref{sec: apx_dists}). We observe that the distribution of direct-surprisal is often right-skewed with many values near zero. While cloze-surprisal shifts the distribution away from zero and approaches a normal distribution. Furthermore, we plot the distribution of GPT2-base direct and cloze surprisal of metaphoric words in VUA at different sentence positions in Figure~\ref{fig: word_pos} (Appendix~\ref{sec: apx_dists}). Again, we observe that cloze-surprisal shifts the distributions away from near zero values. However, against our intuition, the effect of cloze-surprisal is not more impactful in cases where the metaphor is located earlier in the sentence than later. The effect of cloze-surprisal, although positive, is not related to the position of the metaphoric word in the sentence.


\begin{table}
\centering
\small
\begin{tabular}{l | c c c c}
\toprule
\textbf{Model} & \textbf{VUA-r} & \textbf{VUA-d} & \textbf{Lai} & \textbf{GPT-4o}  \\
\midrule
GPT2-base        & \posneg{+4.9}  & \posneg{+4.0}  & \posneg{-7.7} & \posneg{-16.7} \\
GPT2-med      & \posneg{+6.8}  & \posneg{+4.5}  & \posneg{-11.8} & \posneg{-26.9} \\
GPT2-large       & \posneg{+6.2}  & \posneg{+8.2}  & \posneg{-14.2} & \posneg{-25.1} \\
GPT2-xl      & \posneg{+6.3}  & \posneg{+2.5}  & \posneg{-16.0} & \posneg{-28.1} \\
\arrayrulecolor{black!30}\midrule
\arrayrulecolor{black}
Llama-1B          & \posneg{+7.3}  & \posneg{+8.0}  & \posneg{-13.3} & \posneg{-15.7} \\
Llama-1B-It. & \posneg{-3.8} & \posneg{-1.2} & \posneg{-2.4} & \posneg{-10.3} \\
Llama-3B          & \posneg{-0.5} & \posneg{+5.6}  & \posneg{+0.3}  & \posneg{+0.7}  \\
Llama-3B-It. & \posneg{-11.1} & \posneg{-2.3} & \posneg{+4.7}  & \posneg{+6.5}  \\
Llama-8B          & \posneg{+1.9}  & \posneg{+2.7}  & \posneg{+2.5}  & \posneg{+12.7}  \\
Llama-8B-It. & \posneg{-7.8} & \posneg{-0.2} & \posneg{+6.0}  & \posneg{+10.9}  \\
\arrayrulecolor{black!30}\midrule
\arrayrulecolor{black}
Qwen-0.5B         & \posneg{-2.1} & \posneg{+2.1}  & \posneg{-2.5} & \posneg{+6.2}  \\
Qwen-0.5B-It.   & \posneg{-4.5} & \posneg{-1.6} & \posneg{-3.9} & \posneg{+7.0}  \\
Qwen-7B           & \posneg{+3.7}  & \posneg{+5.6}  & \posneg{+2.4}  & \posneg{+12.8}  \\
Qwen-7B-It.  & \posneg{-6.9} & \posneg{-1.6} & \posneg{+9.1}  & \posneg{+8.0}  \\
Qwen-14B          & \posneg{+5.8}  & \posneg{+7.8}  & \posneg{+3.5}  & \posneg{+6.9}  \\
Qwen-14B-It. & \posneg{-4.4} & \posneg{+0.7}  & \posneg{+4.6}  & \posneg{+12.8}  \\
\bottomrule
\end{tabular}
\caption{Cloze-surprisal \% gains (over direct-surprisal from the same model) in Rank-biserial's correlation estimates between surprisal and novelty scores/labels in the four datasets: \textbf{VUA-r}atings, \textbf{VUA-d}ictionary, \textbf{Lai}2009 and \textbf{GPT-4o}-metaphors.}
\label{tab: results_cloze_gain}
\end{table}

\begin{table}[!htp]
\centering
\small
\begin{tabular}{l|c|cc|cc}
\toprule
\textbf{Genre} &    & \multicolumn{2}{c}{\textbf{VUA-ratings}} & \multicolumn{2}{|c}{\textbf{VUA-dictionary}} \\
\midrule
& ppl. & $r_b$ & Nov. \% & $r_b$ &  Nov. \%  \\
\midrule
Fiction     &  108 	&	 .693 &	 2.97		&	 .478 &	 3.56	 	 \\
News 	    &  89	   &	 .653 & 2.80	 		&	 .413 &	 3.88	 \\
Academic 	&  73	 &	 .588 	&      1.85	 	&	 .413 	 &	 1.85 \\
Conversation 	&  134	 	&	 .482  & 1.41		&	 .800 &	 0.62 	 \\
\arrayrulecolor{black!30}\midrule
\arrayrulecolor{black}
All & 96  & .638  & 2.33  & .581 & 2.70 \\
\bottomrule
\end{tabular}
\caption{GPT2-base perplexity, metaphor density, and GPT2-base surprisal correlation with metaphor novelty of VUA-ratings and VUA-dictionary per genre.}
\label{tab: genre_measures}
\end{table}

\subsection{Genre Splits}
In Table~\ref{tab: genre_measures}, we report the rank-biserial correlations between GPT2-base surprisals and binary novelty labels of VUA-ratings and VUA-dictionary on each genre split separately. We also report the share of novel metaphors within each genre to illustrate the variance in metaphor density across genres. Additionally, we report the perplexity of GPT2-base on each genre split separately.

For VUA-ratings, we find a positive relation between genres' metaphor density and the correlation of surprisal with novelty. We also observe a positive relation between perplexity and surprisal-novelty correlations that is only violated on the conversation split. On the other hand, for VUA-dictionary, the relation between metaphor density and surprisal-novelty correlation is not preserved. While the perplexity positive relation with surprisal-novelty correlation is more concrete. Interestingly, in contrast to VUA-ratings, the dictionary-based annotations of the conversation split correlate strongly with surprisal values despite its tiny number of annotated novel metaphors (11 out of 1774 metaphors) and high perplexity (134).

\section{Discussion}
\label{sec: discussion}

\paragraph{Is surprisal a good metric for metaphor novelty?}
Across multiple novelty annotation setups—human ratings, dictionary-based binary labels, experimental contrastive items designed by experts,
and LLM-generated conventional vs.\ novel senses—we find consistently moderate associations between LM surprisal and metaphor novelty (best: $r=.49, \rho=.50, r_b=.69, \mathrm{AUC}=.84$). While these results are not directly comparable to prior work due to differences in datasets and task formulations, their magnitudes are broadly in line with reported surprisal--behavior associations in acceptability judgments~\cite{tjuatja-etal-2025-goes} and reading times~\cite{oh-schuler-2023-transformer}. The closest point of reference in the metaphor novelty literature is \cite{parde_corpus_2018}, who report $r=.44$ when predicting novelty scores from a wide range of linguistic features. They additionally report moderate inter-annotator agreement, underscoring that novelty itself is challenging to measure reliably by humans.
At the same time, surprisal has clear theoretical limits as a standalone predictor: interpreting a metaphor involves semantic integration and cross-domain mapping, not only predictability. We therefore expect surprisal to be most informative when combined with measures that more directly target novel interpretations and domain mappings, especially for highly creative (novel) metaphors.

\paragraph{Model Sizes \& Dataset Types:} Our results introduce a great opportunity to further understand the underlying factors of opposing negative and positive effects of model sizes on correlations with psycholinguistic and cognitive features. The negative and positive effects are present together in our study, and clearly contrasted by the type of metaphor novelty dataset under experiment.

The negative effect of model size observed in corpus-based datasets mirrors the inverse scaling effect reported in reading time studies~\cite{oh-schuler-2023-surprisal,WILCOX2025104650}. Recently, \cite{oh-etal-2024-frequency} argued that this effect is largely driven by word frequency, with larger models assigning increasingly non-human-like expectations to rare words. Since corpus-based novelty scores are often correlated with word frequency~\cite{do-dinh-etal-2018-weeding,reimann-scheffler-2024-metaphor}, frequency can be an underlying factor for this negative scaling effect in our study.
By contrast, the synthetic datasets control for lexical identity and frequency: conventional and novel senses are elicited for the same set of words, and stimuli are constructed to cleanly separate the two senses. Under these controls, scaling improves alignment with novelty labels, agreeing with reports of positive scaling effects in other behavioural settings such as acceptability judgments and other reading time studies~\cite{tjuatja-etal-2025-goes,wilcox-etal-2023-language}.
Overall, we see our results as evidence 
that these diverging scaling effects are due to the nature of the dataset types. While corpus-based datasets reflect metaphor novelty mainly through lexical properties such as word frequency, synthetic datasets more directly isolate the conventional–novel distinction by controlling for lexical properties confounds.

\paragraph{Cloze-surprisal and Instruction-tuning:} 
Our cloze-surprisal approach improves correlation across many model variants. In Figures~\ref{fig: scores_surp_hist} and~\ref{fig: word_pos}, cloze-surprisal is found to be shifting the surprisal values away from near-zero values and pushing towards a normal distribution.\\
In examples from Table \ref{tab: data_examples}, cloze-surprisal often raises the surprisal of the metaphorical word, consistent with the intuition that the full context is needed to process the metaphor and realise its true predictability. Moreover, in cases where the metaphorical word begins a sentence (e.g. example 14), direct surprisal is naturally high—regardless of novelty—whereas cloze-surprisal successfully moderates such inflated values.\\
Instruction-tuning—despite its goal of aligning model outputs with human intent—does not enhance the human-likeness of surprisal. In fact, similar to prior findings~\cite{kuribayashi-etal-2024-psychometric}, instruction-tuned models tend to reduce alignment between predicted probabilities and human annotation scores.

\paragraph{Genre Effects: } Our results show a significant effect of sentences' genre on surprisal correlation with metaphor novelty. We suspect the difference in genre's novel metaphor density can be an underlying factor. Also, our observations on perplexity relations to the correlations trigger the possibility that the amount of pretraining data contributing to each genre can significantly affect LM-based methods of detecting novel metaphors.

\section{Conclusion}

We have studied the distinction between conventional and novel metaphors and systematically investigated surprisal computed with LMs as a metric for metaphor novelty.
In general, our experiments show some potential for surprisal in predicting aspects of linguistic creativity, but also call for novel
measures and datasets that provide systematic annotations of metaphor novelty across genres and across corpus-based and experimental settings.

\section*{Limitations}

We acknowledge a couple of limitations in this work. First, the scarcity of high-quality metaphor novelty annotations in existing literature constrains both coverage and generalizability. Second, we rely on pretrained language models whose training data and processes are not fully disclosed. Additionally, although we carefully analyse the effects of model architecture, size, and domain (genre), future work could adopt mixed-effects models to test the interaction of these variables.

\section*{Acknowledgments}
This research has been funded by the Deutsche Forschungsgemeinschaft (DFG, German Research Foundation) – CRC-1646, project number 512393437, project A05.\\
We acknowledge the anonymous reviewers and area chairs for their valuable comments and feedback. Furthermore, we thank Özge Alaçam, Annett Jorschick, Vicky Tzuyin Lai and Tiago Pimentel for their cooperation and responsiveness to our inquiries.

\bibliography{anthology_used}

\appendix

\section{Appendix}

\subsection{Statistics and Examples}
\label{sec: ap_stats_examples}

In Tables~\ref{tab: data_stats} and \ref{tab: data_stats2}, we describe in numbers the four datasets under study. In Table~\ref{tab: data_examples}, we list 16 examples from the VUA datasets, 6 examples from the LAI2009 dataset, and 4 examples from the GPT-4o-metaphors dataset.

\begin{table*}[!htp]
\centering
\small
\begin{tabular}{lcc|cc|c}
\toprule
& & & \multicolumn{2}{c}{\textbf{VUA-ratings}} & \multicolumn{1}{|c} {\textbf{VUA-dictionary}} \\
\midrule
Genre & \# Metaphors & L$_{sent}$  & Novelty Score & \# Novel & \# Novel \\
&  & mean | std.  & mean | std. & $>=0.5$  & \\
\midrule
Fiction & 3170 & 26.0 | 16.5 &  -.005 | .271 & 94  & 113 \\
News & 4712 & 29.9 | 14.2 & .000 | .257  & 132 & 183 \\
Academic & 5499 & 34.9 | 16.0  & .003 | .239 & 102 & 102 \\ 
Conversation & 1774 & 17.5 | 15.9   & -.000 | .236 & 25  & 11 \\
\midrule
All & 15155 & 29.4 | 16.5  & .000 | .251 & 353  & 409 \\
\bottomrule
\end{tabular}
\caption{Distributions and statistics of the datasets under study. \# Met. is the number of metaphor words, Score$_{nov}$ is the BWS novelty scores, \# Nov. is the number of novel metaphors (Score$_{nov}$ >= 0.5). L$_{sent}$ is the length of sentences in words. We report the number of metaphors used in the sense of Nouns (\textit{N}), Verbs (\textit{V}) or Adjectives (\textit{Adj.}) in the synthetic datasets.}
\label{tab: data_stats}
\end{table*}

\begin{table}[!htp]
\centering
\small
\begin{tabular}{l|cc|cc}
\toprule
& \multicolumn{2}{c|} {\textbf{Lai2009}} & \multicolumn{2}{c} {\textbf{GPT-4o-metaphors}} \\
\midrule
Label & \# & L$_{sent}$
& \# & L$_{sent}$
\\
&  & mean | std.
&  & mean | std.
\\
\midrule
Conventional & 104 & 6.7 | 1.5
& 100 & 10.1 | 1.9
\\
Novel & 104 & 6.7 | 1.4
& 100 & 12.7 | 2.3
\\
\midrule
All & 208 & 6.7 | 1.5
& 200 & 11.4 | 2.5
\\

\bottomrule
\end{tabular}
\caption{Distributions and statistics of the datasets under study. \# Met. is the number of metaphor words, Score$_{nov}$ is the BWS novelty scores, \# Nov. is the number of novel metaphors (Score$_{nov}$ >= 0.5). L$_{sent}$ is the length of sentences in words. We report the number of metaphors used in the sense of Nouns (\textit{N}), Verbs (\textit{V}) or Adjectives (\textit{Adj.}) in the synthetic datasets.}
\label{tab: data_stats2}
\end{table}

\begin{table*}[!htp]
\centering
\small
\begin{tabular}{p{9cm}|c|c|c|c}
\toprule

\multicolumn{5}{c}{\textbf{VUA datasets}} \\
\midrule
\multicolumn{1}{c|}{Sentence} & ratings & dictionary & direct & cloze \\
\midrule

{\fontsize{7pt}{9pt}\selectfont \textbf{1. } ‘ Tell him I am very sorry, but I must \textbf{fill} the quota. ’} &
-0.441 &
conventional &
9.043 &
7.317
\\
{\fontsize{7pt}{9pt}\selectfont \textbf{2. } Adam might have escaped the file memories for years, \textbf{suppressed} them and jerked violently <14> by those events.} &
0.531 &
novel &
14.23 &
14.69
\\ \cmidrule(lr){1-5}

{\fontsize{7pt}{9pt}\selectfont \textbf{3. } It was an excitement that <11> and I had long dreamed of that scatter of tiny, \textbf{magically} named islands strewn across one third of a globe.} &
0.278 &
conventional &
10.17 &
16.08
\\
{\fontsize{7pt}{9pt}\selectfont \textbf{4. } The seemingly random and <11> designed to disguise a boat's shape from the \textbf{prying} eyes of U-Boat captains, so it <10> in the Bahamas.} &
0.588 &
conventional &
7.903 &
12.46
\\ \cmidrule(lr){1-5}

{\fontsize{7pt}{9pt}\selectfont \textbf{5. } One Mr Clarke can not duck away from if he wants to \textbf{avoid} a second Winter of Discontent} &
 -0.094 &
 conventional &
3.165 &
7.454
\\
{\fontsize{7pt}{9pt}\selectfont \textbf{6. } This was conveniently \textbf{encapsulated} in the first try.} &
0.500 &
conventional &
7.918 &
14.54
\\ \cmidrule(lr){1-5}

{\fontsize{7pt}{9pt}\selectfont \textbf{7. } Thrusts of resistance ( mass demonstrations, resignations, tax rebellions, etc ) would come in \textbf{crests}.} &
0.382 &
novel &
12.46 &
16.84
\\
{\fontsize{7pt}{9pt}\selectfont \textbf{8. } Travel: A \textbf{pilgrimage} sans progress Elisabeth de Stroumillo potters round Poitou} &
0.514 &
conventional &
8.188 &
14.81
\\ \cmidrule(lr){1-5}

{\fontsize{7pt}{9pt}\selectfont \textbf{9. } The Tehuana dress is by no means the most decorative variant or the \textbf{closest} to pre-Hispanic forms of clothing.} &
 -0.194 &
 conventional &
6.466 &
11.92
\\
{\fontsize{7pt}{9pt}\selectfont \textbf{10. } Interwoven with these images are subtler references to the \textbf{metaphorical} borderlines which separate Latin American <5> and North America.} &
0.529 &
conventional &
10.16 &
12.57
\\ \cmidrule(lr){1-5}

{\fontsize{7pt}{9pt}\selectfont \textbf{11. } This is often linked with a supposed denunciatory effect — the idea that the mandatory life sentence \textbf{denounces} murder as emphatically as possible <18> this crime.} &
0.294 &
conventional &
11.02 &
12.61
\\
{\fontsize{7pt}{9pt}\selectfont \textbf{12. } He certainly held deep convictions as to the <9>, but at least a part of his apparent \textbf{hostility} was assumed for the occasion, a hard <7> in the end.} &
0.514 &
conventional &
5.662 &
9.781
\\ \cmidrule(lr){1-5}

{\fontsize{7pt}{9pt}\selectfont \textbf{13. } Me dad said he's had enough Well, we were debating whether to \textbf{give} it to you or not.} &
 -0.633 &
 conventional &
3.752 &
8.038
\\
{\fontsize{7pt}{9pt}\selectfont \textbf{14.} \textbf{Struggled} with it a little} &
0.552 &
conventional &
17.13 &
14.92
\\ \cmidrule(lr){1-5}

{\fontsize{7pt}{9pt}\selectfont \textbf{15. }That's an old \textbf{trick}.} &
0.310 &
conventional &
4.013 &
11.92
\\
{\fontsize{7pt}{9pt}\selectfont \textbf{16. }Can you \textbf{sort} erm, madame out?} &
0.567 &
conventional &
8.820 &
9.591
\\

\midrule
\multicolumn{4}{c}{\textbf{LAI2009}} \\
\midrule

\multicolumn{1}{c|}{Sentence} & \multicolumn{2}{c|}{label} & direct & cloze \\
\midrule

{\fontsize{7pt}{9pt}\selectfont \textbf{17. }Upon hearing the news my spirits \textbf{sank}} &
\multicolumn{2}{c|}{conventional} &
4.358 &
11.82
\\
{\fontsize{7pt}{9pt}\selectfont \textbf{18. }Upon having the data my prediction \textbf{sank}} &
\multicolumn{2}{c|}{novel} &
10.28 &
13.27
\\ \cmidrule(lr){1-5}

{\fontsize{7pt}{9pt}\selectfont \textbf{19. }Those chess players are prepared for \textbf{battle}} &
\multicolumn{2}{c|}{conventional} &
5.373 &
12.32
\\
{\fontsize{7pt}{9pt}\selectfont \textbf{20. }Those plastic surgeons are prepared for \textbf{battle}} &
\multicolumn{2}{c|}{novel} &
6.691 &
13.28
\\ \cmidrule(lr){1-5}

{\fontsize{7pt}{9pt}\selectfont \textbf{21. }His mental condition remains \textbf{fragile}} &
\multicolumn{2}{c|}{conventional} &
7.251 &
13.34
\\
{\fontsize{7pt}{9pt}\selectfont \textbf{22. }His website popularity remains \textbf{fragile}} &
\multicolumn{2}{c|}{novel} &
9.538 &
13.95
\\

\midrule
\multicolumn{4}{c}{\textbf{GPT-4o-metaphors}} \\
\midrule

{\fontsize{7pt}{9pt}\selectfont \textbf{23. }Her family was her emotional \textbf{anchor} during the crisis.} &
\multicolumn{2}{c|}{conventional} &
4.322 &
14.12
\\
{\fontsize{7pt}{9pt}\selectfont \textbf{24. }The smell of coffee became an \textbf{anchor} to mornings that no longer came.} &
\multicolumn{2}{c|}{novel} &
7.518 &
12.42
\\ \cmidrule(lr){1-5}

{\fontsize{7pt}{9pt}\selectfont \textbf{25. }The software helps users \textbf{navigate} complex legal documents.} &
\multicolumn{2}{c|}{conventional} &
3.830 &
5.575
\\
{\fontsize{7pt}{9pt}\selectfont \textbf{26. }He tried to \textbf{navigate} the silence like a sailor without stars.} &
\multicolumn{2}{c|}{novel} &
8.041 &
8.402
\\

\bottomrule
\end{tabular}

\caption{Examples from each dataset. The metaphor word is in \textbf{boldface} within sentences. For simpler presentations, we remove some words from long sentences and replace them with a tag of the number of words removed, e.g. <11>. \textbf{rating} is the \cite{do-dinh-etal-2018-weeding} score, and \textbf{dictionary} is the \cite{reimann-scheffler-2024-metaphor} annotation. \textbf{direct} is the GPT2-Base recorded surprisal, and cloze is the GPT2-Base cloze-surprisal. Paired examples are picked randomly for VUA datasets from the same genres to illustrate the differences between conventional and novel instances according to ``ratings''. Sentences 1-4 are from Fiction, 5-8 from News, 9-12 from Academic and 13-16 from Conversation. Also obviously, \cite{do-dinh-etal-2018-weeding} ratings and \cite{reimann-scheffler-2024-metaphor} annotations do not agree in many cases.}
\label{tab: data_examples}
\end{table*}

\subsection{GPT-4o-metaphors Construction}
\label{sec: apx_prompt}

To generate the sentences in GPT-4o-metaphors, we prompt GPT-4o once for each word in the dataset. As we planned for 10 different words, we prompted the model 10 consecutive times to construct this dataset. \textcolor{blue}{\texttt{``I am curating a dataset to be used in a study about metaphoric knowledge in pretrained language models PLMs e.g. GPT-2. The dataset should consist of sentences that correspond to target nouns/verbs. 
For each target noun/verb, there should be 10 sentences. 
The target noun/verb should be used 10 times using conventional metaphoric meanings of the target noun/verb, and 10 times using novel metaphoric meanings of the target noun/verb. 
Please suggest a target noun, and generate 20 sentences following the requirements above: 10 sentences with Conventional Metaphor usages, and 10 sentences with Novel Metaphor usages.''}}

\subsection{Models Details}
\label{sec: apx_models}

All the LMs used in this study are based on the HuggingFace HF models hub. Hence, we list here the exact HF models' IDs that we used in our experiments. We use default HF parameters when forwarding the inputs through the model layers. For models meta-data and parameters, please refer to the models' cards on HF using the links below:

\begin{enumerate}
    \item \href{https://huggingface.co/openai-community/gpt2}{openai-community/gpt2}
    \item \href{https://huggingface.co/openai-community/gpt2-medium}{openai-community/gpt2-medium}
    \item \href{https://huggingface.co/openai-community/gpt2-large}{openai-community/gpt2-large}
    \item \href{https://huggingface.co/openai-community/gpt2-xl}{openai-community/gpt2-xl}
    
    \item \href{https://huggingface.co/meta-llama/Llama-3.2-1B}{meta-llama/Llama-3.2-1B}
    \item \href{https://huggingface.co/meta-llama/Llama-3.2-1B-Instruct}{meta-llama/Llama-3.2-1B-Instruct}
    \item \href{https://huggingface.co/meta-llama/Llama-3.2-3B}{meta-llama/Llama-3.2-3B}
    \item \href{https://huggingface.co/meta-llama/Llama-3.2-3B-Instruct}{meta-llama/Llama-3.2-3B-Instruct}
    \item \href{https://huggingface.co/meta-llama/Llama-3.1-8B}{meta-llama/Llama-3.1-8B}
    \item \href{https://huggingface.co/meta-llama/Llama-3.1-8B-Instruct}{meta-llama/Llama-3.1-8B-Instruct}

    \item \href{https://huggingface.co/Qwen/Qwen2.5-0.5B}{Qwen/Qwen2.5-0.5B}
    \item \href{https://huggingface.co/Qwen/Qwen2.5-0.5B-Instruct}{Qwen/Qwen2.5-0.5B-Instruct}
    \item \href{https://huggingface.co/Qwen/Qwen2.5-7B}{Qwen/Qwen2.5-7B}
    \item \href{https://huggingface.co/Qwen/Qwen2.5-7B-Instruct}{Qwen/Qwen2.5-7B-Instruct}
    \item \href{https://huggingface.co/Qwen/Qwen2.5-14B}{Qwen/Qwen2.5-14B}
    \item \href{https://huggingface.co/Qwen/Qwen2.5-14B-Instruct}{Qwen/Qwen2.5-14B-Instruct}
\end{enumerate}

\subsection{Use of AI Assistants}
AI assistants were used during manuscript preparation only for limited linguistic editing to improve clarity and style, and for writing auxiliary code (e.g., for visualisations). They were not used for scientific reasoning, evaluation decisions, or interpretation of results; all analyses and conclusions were drawn by the authors.

\subsection{Scientific Artifacts}

VUA datasets are publicly available datasets and intended for scientific research, and we follow this purpose in this study. Lai2009 dataset has copyright regulations; the author of this dataset approved our usage of the dataset, and she promised to share the data with anyone interested in reproducing our experiments. We double-checked that the datasets and prompts do not contain any personal data.

\subsection{Distributions Visualisations}
\label{sec: apx_dists}

\begin{figure*}[t]
  \centering
  \includegraphics[width=1\linewidth]{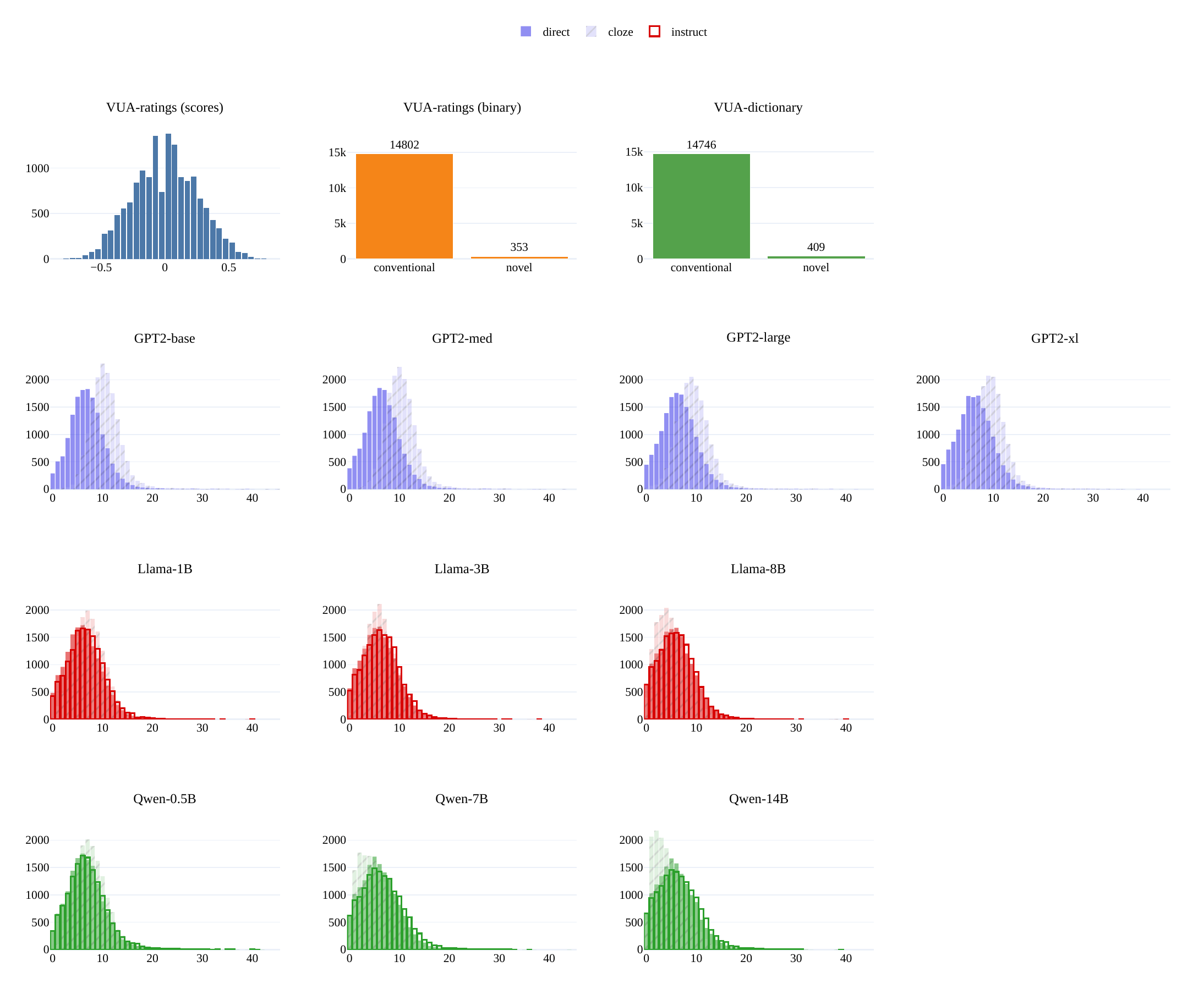}
  \caption{Histograms of VUA metaphor novelty annotations, and GPT2-base direct surprisal (dark colour) of content metaphoric words (15,155) across 10 LMs. Additionally, cloze-surprisal of the same models is plotted in (dashed) light colour, and surprisal from instruct-tuned variants (whenever applicable) is plotted with borderlines.}
  \label{fig: scores_surp_hist}
\end{figure*}

\begin{figure*}[t]
  \centering
  \includegraphics[width=1\linewidth]{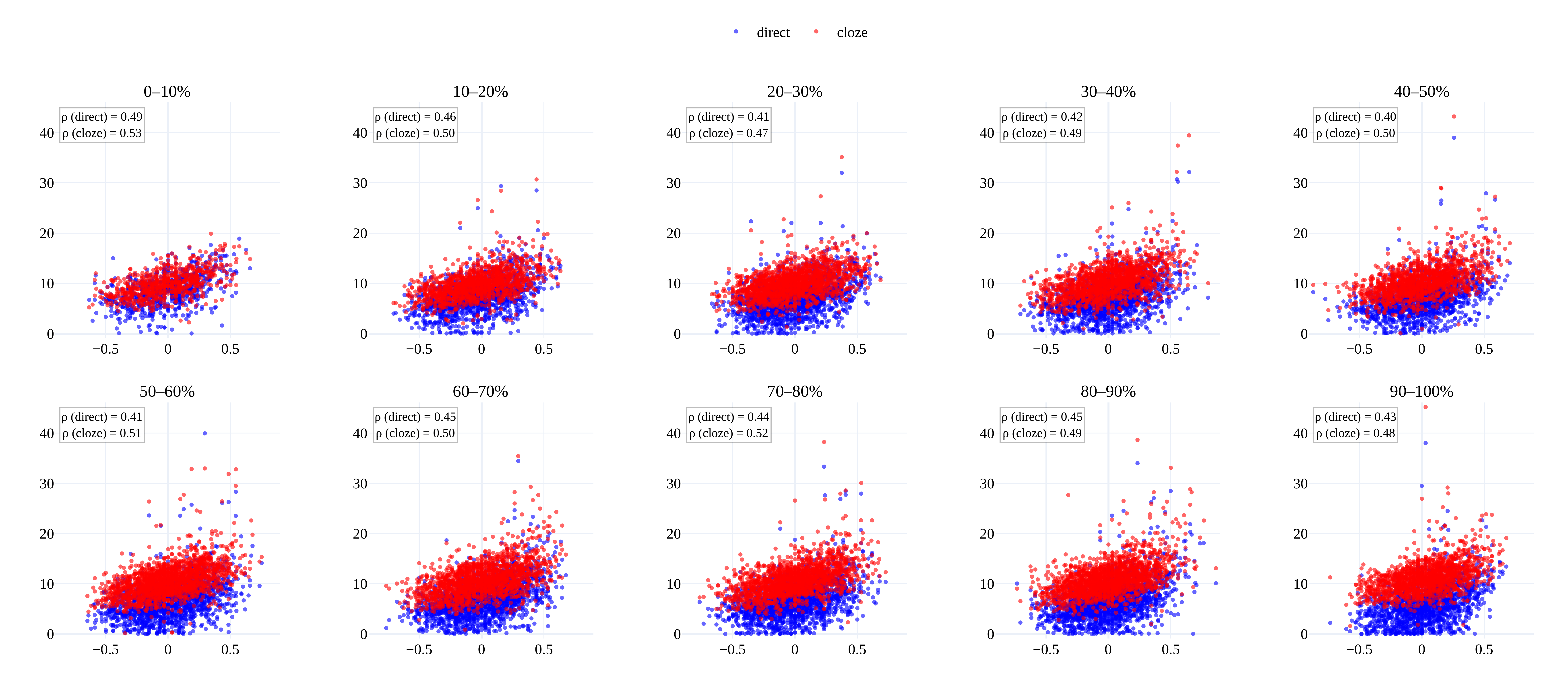}
  \caption{Scatter plots between metaphor novelty scores of VUA-ratings on the x-axis; and direct surprisal (blue) and cloze surprisal (red) (from GPT2-base) on the y-axis at different relative positions of metaphoric words in sentences. 0-10\% means metaphoric words located at the first $1/10 * Length$ of the sentence.}
  \label{fig: word_pos}
\end{figure*}

In Figure~\ref{fig: scores_surp_hist}, we plot the distributions of novelty scores/labels and surprisal of the VUA datasets.\\
In Figure~\ref{fig: word_pos}, we plot scatter plots between VUA-ratings scores and GPT2-base surprisal from both direct and cloze methods, grouping instances based on the location of the metaphoric word within the sentence.

\end{document}